\documentclass[10pt,twocolumn,letterpaper]{article}

\usepackage{cvpr}              

\definecolor{cvprblue}{rgb}{0.21,0.49,0.74}
\usepackage[pagebackref,breaklinks,colorlinks,allcolors=cvprblue]{hyperref}

\def\paperID{*****} 
\def\confName{CVPR}
\def\confYear{2025}

\title{TeSG: Textual Semantic Guidance for Infrared and Visible Image Fusion}

\author{Mingrui~Zhu\\
Xidian University\\
{\tt\small mrzhu@xidian.edu.cn}
\and
Xiru~Chen\\
Xidian University\\
{\tt\small xrchen@stu.xidian.edu.cn}
\and
Xin Wei\\
Xidian University\\
{\tt\small weixin@xidian.edu.cn}
\and
Nannan~Wang\\
Xidian University\\
{\tt\small nnwang@xidian.edu.cn}
\and
Xinbo~Gao\\
Chongqing University of Post and Telecommunications\\
{\tt\small gaoxb@cqupt.edu.cn}
}

\usepackage{multirow}

\begin{document}

\maketitle

\begin{abstract}
  Infrared and visible image fusion (IVF) aims to combine complementary information from both image modalities, producing more informative and comprehensive outputs. Recently, text-guided IVF has shown great potential due to its flexibility and versatility. However, the effective integration and utilization of textual semantic information remains insufficiently studied. To tackle these challenges, we introduce textual semantics at two levels: the mask semantic level and the text semantic level, both derived from textual descriptions extracted by large Vision-Language Models (VLMs). Building on this, we propose \textbf{Te}xtual \textbf{S}emantic \textbf{G}uidance for infrared and visible image fusion, termed TeSG, which guides the image synthesis process in a way that is optimized for downstream tasks such as detection and segmentation. Specifically, TeSG consists of three core components: a Semantic Information Generator (SIG), a Mask-Guided Cross-Attention (MGCA) module, and a Text-Driven Attentional Fusion (TDAF) module. The SIG generates mask and text semantics based on textual descriptions. The MGCA module performs initial attention-based fusion of visual features from both infrared and visible images, guided by mask semantics. Finally, the TDAF module refines the fusion process with gated attention driven by text semantics. Extensive experiments demonstrate the competitiveness of our approach, particularly in terms of performance on downstream tasks, compared to existing state-of-the-art methods.
\end{abstract}    
\section{Introduction}
\label{sec:intro}

\begin{figure}
\centering
\includegraphics[width=1.0\columnwidth]{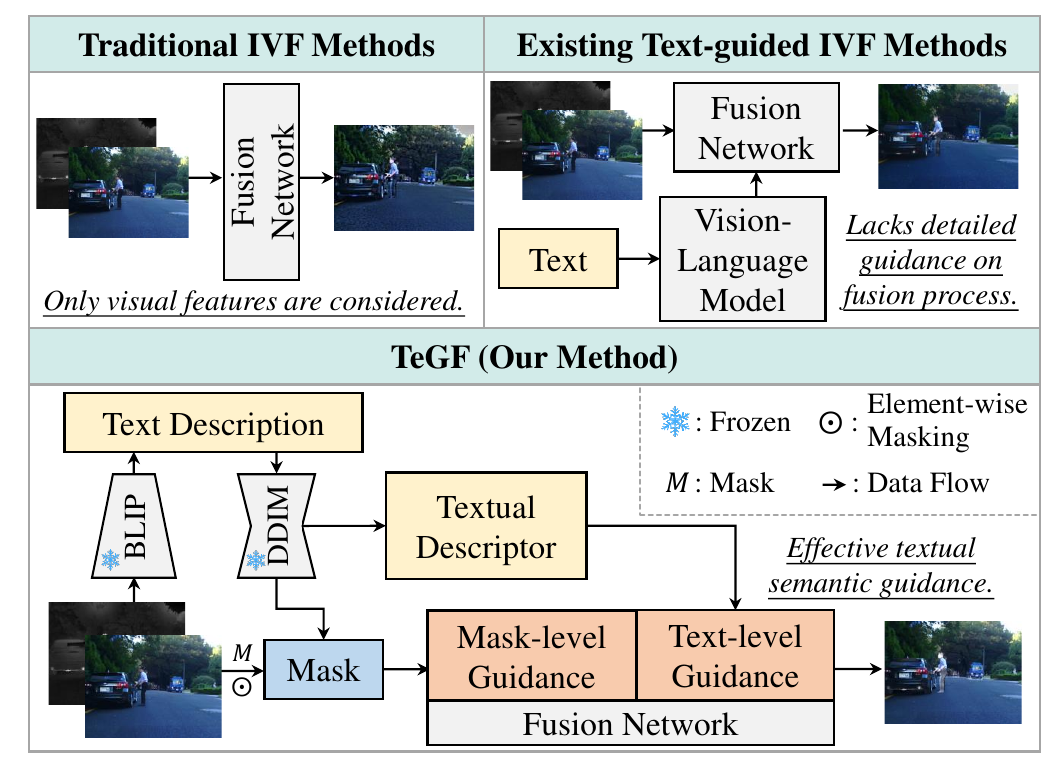}
\caption{Comparison of existing IVF methods with our TeSG. Traditional methods focus only on low-level image features, while current text-guided IVF methods often overlook intrinsic details during the fusion process. In contrast, our approach combines mask-level, text-level, and image-level features to effectively guide the fusion process.}
\label{Intro:Fig.1}
\end{figure}

Due to hardware constraints and the inherent limitations of imaging technologies, single-modal sensors often fail to capture all the relevant information in complex scenes. Infrared images excel in low-light or challenging conditions, offering strong thermal radiation detection, but they lack fine details. On the other hand, visible images provide rich texture details but struggle in low-light or low-contrast environments. As a result, neither modality alone can offer a comprehensive and accurate description of a scene~\cite{ma2019infrared,zhang2021image}. Infrared and visible image fusion (IVF) is a crucial technique for leveraging the complementary strengths of both. The primary goal of IVF is to preserve the thermal characteristics of infrared images while retaining the texture details of visible images, resulting in high-quality fused images. These fused images, with enhanced scene representation, find wide-ranging applications in object detection~\cite{davis2007background,han2013fast}, scene understanding~\cite{xu2016multimodal}, remote sensing~\cite{li2017image}, and other vital domains.

Over the past few decades, many studies have tackled the challenges of IVF. Traditional methods rely on techniques like multi-scale transformations~\cite{li2013image,ma2017infrared}, sparse representation~\cite{li2016improved,li2021infrared}, subspace~\cite{cvejic2007region, wang2014fusion}, and saliency detection~\cite{bavirisetti2016two,liu2017infrared} to directly combine multimodal information. While effective to some extent, these methods rely on fixed feature extraction and manually designed fusion rules, which limit their ability to handle complex and diverse scenarios. Recently, deep learning techniques like autoencoders (AE)~\cite{li2018densefuse,zhao2021didfuse,zhang2021sdnet,liu2021learning}, convolutional neural networks (CNN)~\cite{liu2018infrared,xu2020u2fusion,xu2022rfnet}, generative adversarial networks (GAN)~\cite{ma2019fusiongan,ma2020ddcgan,ma2020infrared,ma2020ganmcc,liu2022target}, and transformers~\cite{ma2022swinfusion,wang2022swinfuse} have significantly improved fusion quality through end-to-end learning. However, these methods focus mainly on visual features and often overlook high-level semantic information and scene context. In complex scenarios, this lack of semantic guidance results in reduced semantic consistency and comprehension, leading to suboptimal fusion performance. 

Most recently, with the revolution in large vision-language models (VLMs), text-guided IVF~\cite{yi2024text,zhao2024image,wang2024terf,cheng2025textfusion} has garnered significant attention. TextIF~\cite{yi2024text} introduces a semantic interaction-guided module that leverages text to specify the types of image degradation, enabling degradation-aware fusion. Zhao \textit{et al.}~\cite{zhao2024image} employs Vision-Language Models (VLMs) to generate detailed semantic descriptions of source images, such as image captions and dense captions, to guide and enhance the fusion of visual features. While the integration of textual information has undoubtedly enhanced the flexibility and performance of existing methods, these approaches still have notable limitations. First, the deep semantic relationship between textual and visual features remains insufficiently explored, with a lack of comprehensive cross-modal interaction and fusion. Second, the precise guidance of textual semantics for effective feature alignment is still largely unaddressed.

To address this issue, we introduce textual semantic guidance, termed TeSG, to effectively steer the fusion process in a manner better aligned with higher-level tasks. TeSG utilizes the pre-trained BLIP model~\cite{li2022blip} to automatically extract textual descriptions of images, thereby eliminating the complexity of manually inputting text. The guidance provided by these textual descriptions during the image fusion process occurs in two key ways: first, by generating mask semantics for key targets based on the textual descriptions, enabling precise mask-level guidance for fusing visual features in critical regions; and second, by leveraging text semantics (feature embeddings of textual descriptions) to establish semantic correlations with visual features, facilitating textual semantic-level gated filtering and guidance for visual feature fusion. To achieve these objectives, TeSG incorporates three functional modules: the Semantic Information Generator (SIG), the Mask-Guided Cross-Attention (MGCA) module, and the Text-Driven Attentional Fusion (TDAF) module. In this way, TeSG effectively leverages textual information and provides a more efficient approach to textual semantic guidance compared to existing text-guided IVF methods, as illustrated in Fig.~\ref{Intro:Fig.1}. Our contributions are summarized below:
\begin{itemize}[leftmargin=*]
\item We explore the potential of using textual descriptions of images to guide the image fusion process and propose an effective textual semantic guidance method. This approach facilitates image fusion through both mask-level and semantic-level guidance, without the need for manual text input. 
\item We design three closely related modules to implement textual semantic guidance, including a semantic information generator, a mask-guided cross-attention module, and a text-driven attentional fusion module.
\item Qualitative and quantitative experiments demonstrate that our method yields higher-quality results, featuring more precise fusion edges and greater information gain, underscoring its effectiveness in supporting higher-level tasks.  
\end{itemize}

\begin{figure*}[!t]
  \centering
  \includegraphics[width=0.85\textwidth]{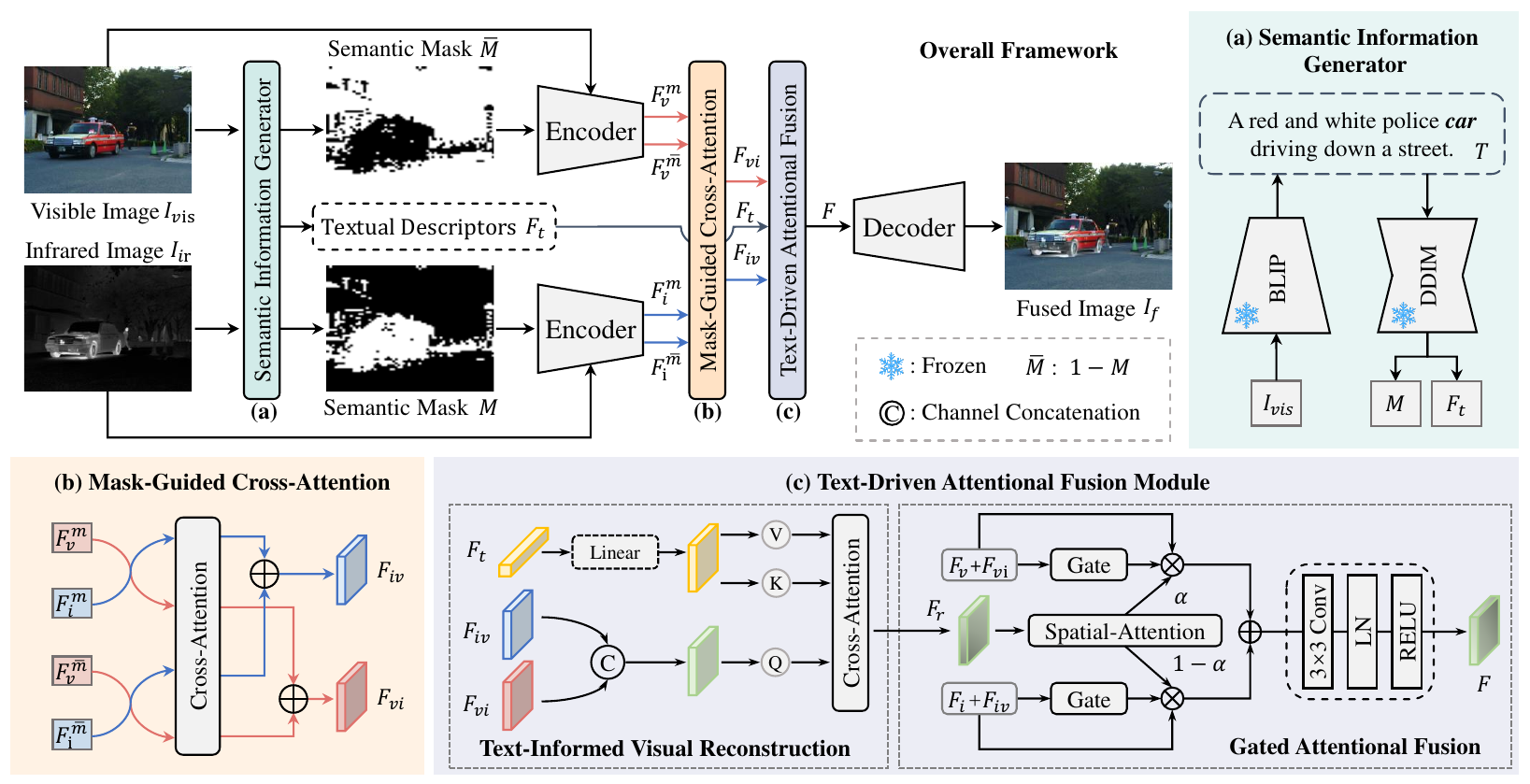}
  \caption{The overall framework of TeSG consists of three primary modules: (a) the Semantic Information Generator (SIG), responsible for generating both mask semantics and text semantics; (b) the Mask-Guided Cross-Attention (MGCA) module, which enables effective cross-modal feature integration through mask-guided attention mechanisms; and (c) the Text-Driven Attentional Fusion (TDAF) module, refining the fusion process by applying gated attention mechanisms driven by the extracted text semantics.}
  \label{Method:Fig.2}
\end{figure*}
\section{Related Work}
\label{sec:Related Work}

\subsection{Infrared and Visible Image Fusion} 


Recent advances in deep learning have significantly promoted the development of infrared and visible image fusion (IVF). Current deep learning-based fusion methods can be further divided into the following four categories. Autoencoder (AE)-based methods~\cite{li2018densefuse,li2021rfn,zhang2021sdnet,liu2021learning} use encoder-decoder frameworks for feature extraction and image reconstruction. For example, DenseFuse~\cite{li2018densefuse} integrates an encoder-decoder network with densely connected convolutional blocks for feature extraction, employing manually designed fusion strategies to integrate features. RFN-Nest~\cite{li2021rfn} proposes an end-to-end residual fusion network, adopting a two-stage training strategy for more effective image fusion. Convolutional Neural Network (CNN)-based methods~\cite{liu2018infrared,xu2020u2fusion,xu2022rfnet,ma2022swinfusion} leverage meticulously designed network architectures and loss functions to extract, integrate, and reconstruct image features. RFNet~\cite{xu2022rfnet} introduces an unsupervised network for multimodal image registration and fusion, utilizing a mutual reinforcement framework to improve fusion performance. Generative Adversarial Network (GAN)-based methods~\cite{ma2019fusiongan,ma2020ddcgan,liu2022target} have gained attention due to their robust ability in unsupervised distribution learning. GANs~\cite{goodfellow2014generative,mirza2014conditional} have demonstrated outstanding performance in infrared and visible image fusion tasks. For instance, FusionGAN~\cite{ma2019fusiongan} employs a generator-discriminator framework to achieve high-quality fusion while effectively preserving fine details. Diffusion-based methods have also emerged as a recent breakthrough in deep learning for image fusion. DDFM~\cite{zhao2023ddfm} utilizes a denoising diffusion probabilistic model for unconditional image generation, excelling in preserving cross-modal information. Similarly, Dif-fusion~\cite{yue2023dif} employs a diffusion model to directly generate the distribution of multi-channel input data, enhancing the aggregation of multi-source information and improving color fidelity.

\subsection{Text-guided IVF} 

Existing methods primarily prioritize the visual quality of fused images, with limited attention given to incorporating semantic information. Some studies attempt to enhance the fused images by integrating downstream tasks, such as object detection~\cite{liu2022target} and semantic segmentation~\cite{tang2022image,liu2023multi}. But these methods are often task-specific and do not facilitate comprehensive interaction or provide fine-grained control over multimodal features, limiting their effectiveness in more general applications. Text-guided methods, on the other hand, directly integrate textual information into IVF tasks. TextIF~\cite{yi2024text} proposes a method that combines an image fusion pipeline with a semantic interaction guidance module to perform perceptual and interactive image fusion tasks on degraded images. Zhao \textit{et al.}~\cite{zhao2024image} introduced VLMs into the image fusion domain, using multi-level textual prompts, such as image captions and dense captions, to guide and enhance the fusion of visual features. TeRF~\cite{wang2024terf} presents a flexible image fusion framework with multiple VLMs, enabling text-driven and region-aware image fusion through the parsing of textual instructions. TextFusion~\cite{cheng2025textfusion} introduces a text-visual association mapping and an affine fusion unit to achieve fine-grained control over the image fusion process. While existing methods have enhanced fusion quality and model flexibility, they still encounter challenges such as high computational complexity and insufficient exploration of deep semantic correlations between textual and visual features. To address these limitations, we propose an efficient text-guided fusion method that facilitates deep multimodal feature integration through dual-level guidance at both the mask and semantic levels, without requiring manually inputted text.
\section{Method}            
\label{sec: Method}

\subsection{Overview}

Given paired samples $\{(I_{vis}, I_{ir})_i\}^N_{i=1}$, where $I_{vis} \in \mathbb{R}^{H \times W \times 3}$ represents samples from a visible image domain and $I_{ir} \in \mathbb{R}^{H \times W \times 1}$ represents samples from a infrared image domain, our goal is to learn a parameterized mapping $I_f = \mathcal{G}(I_{vis}, I_{ir}|\boldsymbol{\theta})$. This mapping enables the generation of a high-fidelity fused image that effectively combines the complementary strengths of both sources. To achieve this, we propose Textual Semantic Guidance (TeSG), a method that effectively directs the fusion process in a way that is better aligned with higher-level tasks. TeSG consists of three key components: the Semantic Information Generator (SIG) module (as described in Sec.~\ref{Sec_SIG}), the Mask-Guided Cross-Attention (MGCA) module (as described in Sec.~\ref{Sec_MGCA}), and the Text-Driven Attentional Fusion (TDAF) module (as described in Sec.~\ref{Sec_TDAF}).

As illustrated in Fig.~\ref{Method:Fig.2}, the aligned visible image $I_{vis}$ and infrared image $I_{ir}$ are first processed by the SIG module, where a frozen BLIP encoder~\cite{li2022blip} extracts textual descriptions from $I_{vis}$. These textual descriptions, along with the input images, are then used in a frozen DDIM sampler~\cite{song2021denoising} to generate mask semantics $M$ and text semantics $F_t$. Following this, both input images and their corresponding mask semantics $M$ are passed through encoders to produce feature representations, including global features (\textit{i.e.}, ${F}_{{v}}$, ${F}_{{i}}$) and local features ({\textit{i.e.}}, $F_{v}^{m}$, $F_{v}^{\overline{m}}$, $F_{i}^{m}$, $F_{i}^{\overline{m}}$). Transformer-based blocks are used for both the encoders and the decoder. These encoded features are then processed by the MGCA module, which facilitates cross-modal feature integration through mask-guided attention mechanisms. The resulting features are passed into the TDAF module, where the text semantics $F_{t}$ play a crucial role in guiding the image fusion process. Additionally, the TDAF module incorporates a dynamic gating mechanism that adjusts fusion weights to ensure an optimal combination of the visible and infrared image features. Finally, the fused image ${I}_{f}$ is reconstructed through the decoder. We adopt commonly used losses for training, including structural similarity loss, gradient loss, intensity loss, and color loss. Detailed formulations are presented in Sec. A of the supplementary material.

\subsection{Semantic Information Generator}  
\label{Sec_SIG}

\begin{figure}
  \centering
  \includegraphics[width=1.0\columnwidth]{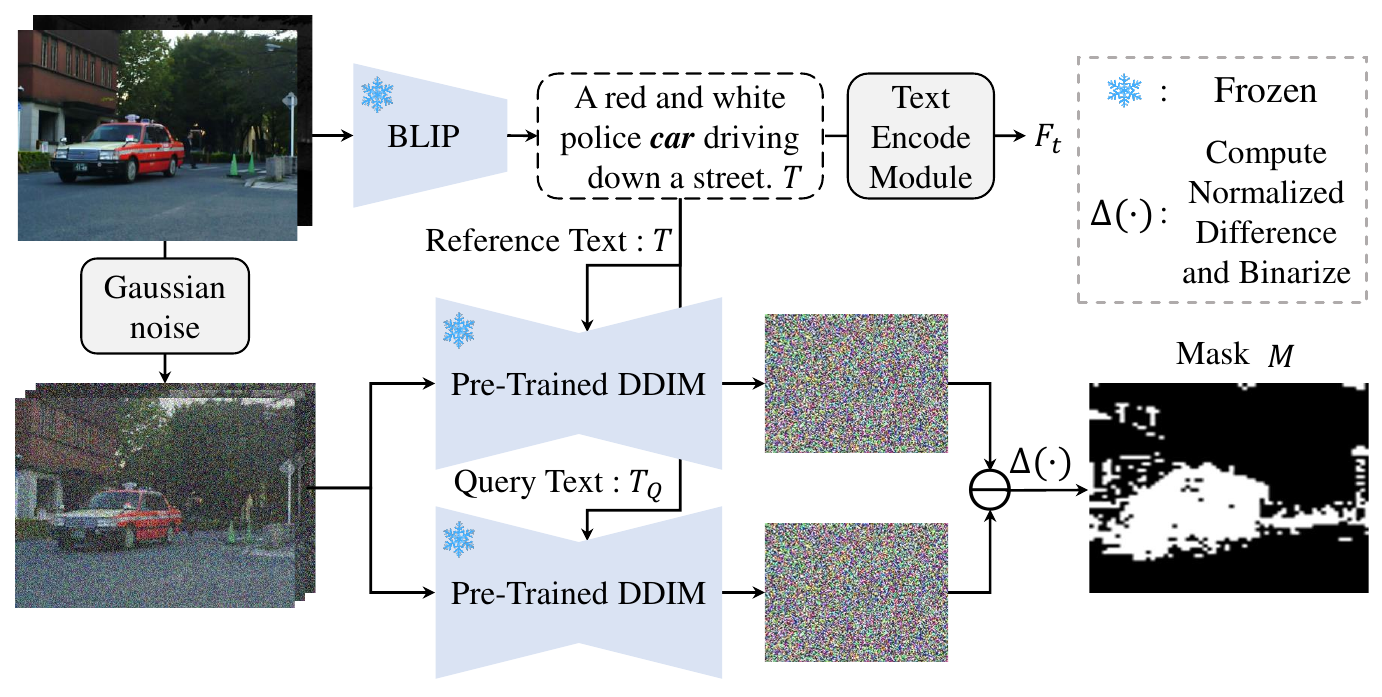}
  \caption{The process of the Semantic Information Generator.}
  \label{Method:Fig.3}
  \end{figure}

\noindent
\textbf{Textual Description Generation.} To generate a contextually accurate textual description of the input images, we leverage the pre-trained BLIP model~\cite{li2022blip}. The visible image ${I}_{{vis}}$ is passed through the frozen BLIP encoder to produce a corresponding textual description ${T}$. This process can be expressed as:
\begin{equation} 
T = \text{BLIP}(I_{vis}),
\end{equation}
where ${I}_{{vis}}$ represents the input visible image, and ${T}$ denotes the generated textual description that captures the key semantic content of the image, as inferred by the BLIP model. Additionally, we modify ${T}$ by removing the keyword ${V^*}$, resulting in the modified textual description ${\widehat{T}}$. Specifically, the keyword ${V^*}$ (\textit{e.g.}, ``car'' in Fig.~\ref{Method:Fig.3}) is defined as the term that corresponds to key objects in high-level tasks, such as object detection and scene segmentation.

\noindent
\textbf{Mask Semantics Generation.} One of the persistent challenges in IVF tasks is managing information redundancy and preventing the loss of critical details during the fusion process. To address these issues, we introduce mask semantics designed to emphasize the complementary features between infrared and visible modalities. Specifically, we propose a noise-based strategy where Gaussian noise is added to the input visible image, $I_{vis}$, and the resulting noisy image is processed under two distinct textual conditions (\textit{i.e.}, ${T}$ and ${\widehat{T}}$) within the diffusion model. This approach enables us to investigate how variations in text influence information within specific regions. The mask is then inferred by evaluating the difference in noise estimation produced by the diffusion model under these two textual conditions. This process can be expressed as:
\begin{equation} 
M_{vis} = \Delta (D_{\theta}(I_{vis}, T) - D_{\theta}(I_{vis},{\widehat{T}})) ,
\end{equation}
where $D_{\theta}$ represents the diffusion model, and $\Delta$ denotes the normalization and binarization of the computed noise difference. To further mitigate the challenge of insufficient information extraction from individual modalities, we extend this approach to both visible and infrared images. Specifically, we generate masks $M_{vis}$ and $M_{ir}$ for the visible image $I_{vis}$ and the infrared image $I_{ir}$, respectively, and then take their union to obtain the final mask semantics ${M}$:
\begin{equation} 
M = M_{vis} \cup M_{ir}.
\end{equation}

\noindent
\textbf{Text Semantics Generation.}
The text encoder of the diffusion model converts the textual description ${T}$ into an embedding vector $F_{t}$, which serves as the text semantics. Overall, the SIG produces both mask-level and text-level semantics to guide the subsequent fusion process.

\subsection{Mask-Guided Cross-Attention}  
\label{Sec_MGCA}

The MGCA module first utilizes mask semantics to decompose the images into key target foreground images and background images. Given the visible image ${I}_{vis}$, the infrared image ${I}_{ir}$, and the corresponding mask semantics ${M}$, the module applies both the foreground semantics ${M}$ and the background semantics $\overline{M}$ to each input image, where $\overline{M}$ is defined as the complementary regions of the foreground mask. This results in four distinct masked images that capture both foreground and background information for each modality. 
These masked images are then passed through an encoder to extract their initial feature representations, denoted as $F_{v}^{m}$, $F_{v}^{\overline{m}}$, $F_{i}^{m}$ and $F_{i}^{\overline{m}}$, where the superscripts $m$ and $\overline{m}$ indicate the foreground and background features, respectively.

Subsequently, we perform feature reconstruction based on cross-attention for the foreground and background features of both visible and infrared images. The foreground and background features of the infrared image, $F_{i}^{m}$ and $F_{i}^{\overline{m}}$, are used to reconstruct the foreground and background features of the visible image, $F_{v}^{m}$ and $F_{v}^{\overline{m}}$, resulting in $F_{vi}^{m}$ and $F_{vi}^{\overline{m}}$:
\begin{equation} 
F_{vi}^{m} = {CA}(F_{v}^{m}, F_{i}^{m}), \quad F_{vi}^{\overline{m}} = {CA}(F_{v}^{\overline{m}}, F_{i}^{\overline{m}});
\end{equation}
conversely, the foreground and background features of the visible image, $F_{v}^{m}$ and $F_{v}^{\overline{m}}$, are used to reconstruct the foreground and background features of the infrared image, $F_{i}^{m}$ and $F_{i}^{\overline{m}}$, resulting in $F_{iv}^{m}$ and $F_{iv}^{\overline{m}}$: 
\begin{equation} 
F_{iv}^{m} = {CA}(F_{i}^{m}, F_{v}^{m}), \quad F_{iv}^{\overline{m}} = {CA}(F_{i}^{\overline{m}}, F_{v}^{\overline{m}}),
\end{equation}
where ${CA}$ denotes the cross-attention mechanism. Finally, the corresponding features are summed element-wise to obtain the fully reconstructed visible and infrared features:
\begin{equation} 
F_{vi} = F_{vi}^{m} + F_{vi}^{\overline{m}}, \quad 
F_{iv} = F_{iv}^{m} + F_{iv}^{\overline{m}}.
\end{equation}

\begin{figure*}[!t]
  \centering
  \includegraphics[width=0.88\textwidth]{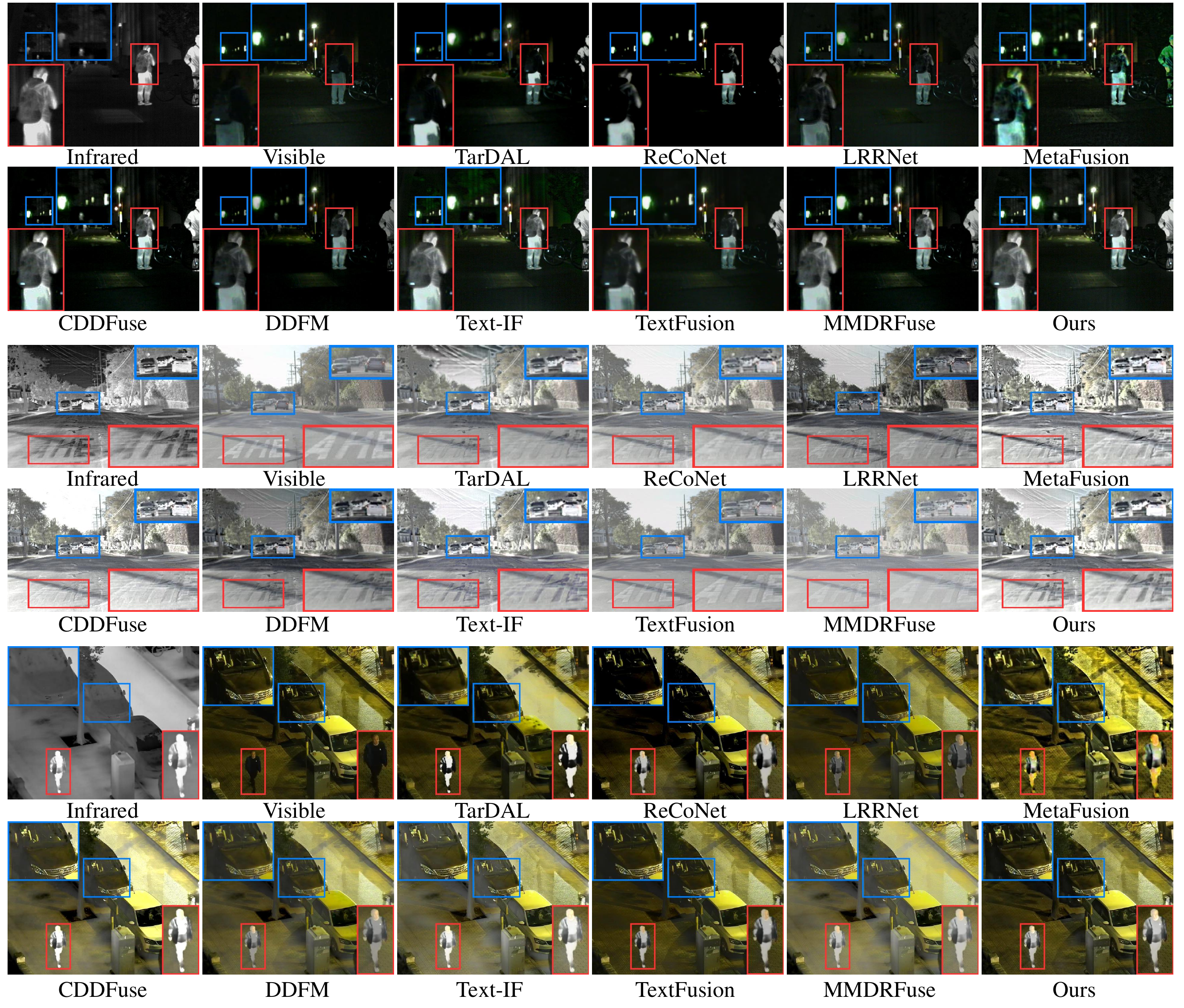}
  \caption{Qualitative comparison of our method against nine state-of-the-art (SOTA) methods. The images, arranged from top to bottom, are sourced from the MSRS, RoadScene, and LLVIP datasets.}
  \label{Experi:Fig.4}
\end{figure*}

\subsection{Text-Driven Attentional Fusion}  
\label{Sec_TDAF}

\noindent
\textbf{Text-Informed Visual Reconstruction.} Given the reconstructed visible and infrared features, $F_{vi}$ and $F_{iv}$, along with the text semantics $F_{t}$, the features $F_{vi}$ and $F_{iv}$ are first concatenated to form the multimodal visual features. The multimodal visual features are then reconstructed using $F_{t}$ through the cross-attention mechanism, enabling effective feature interaction. This process is formally defined as follows:
\begin{equation} 
F_{r} = {CA}({Cat}(F_{vi}, F_{iv}), F_{t}),
\end{equation}
where $Cat(\cdot)$ represents the concatenation of features and ${CA}$ denotes the cross-attention mechanism. Through this process, the module generates the text-informed multimodal visual features $F_{r}$, which integrate text semantic information with multimodal features.

\noindent
\textbf{Gated Attentional Fusion.} To fully leverage the complementary characteristics of infrared and visible features, we introduce a gated mechanism that dynamically weights and fuses the two modalities. First, the reconstructed visible and infrared features, $F_{vi}$ and $F_{iv}$, are added to the initial image features, $F_{v}$ and $F_{i}$, respectively. The resulting features are then processed through a convolutional layer to generate the gating weights, $G_{v}$ and $G_{i}$. The gating weights are computed as follows:
\begin{equation}
\begin{aligned}
G_{v} &= \sigma (W_{v} \cdot (F_{v} + F_{vi})), \\
G_{i} &= \sigma (W_{i} \cdot (F_{i} + F_{iv})),
\end{aligned}
\end{equation}
where $\sigma$ denotes the Sigmoid activation function, and $W_{v}$ and $W_{i}$ represent the weight matrices of the convolutional layers for the visible and infrared features, respectively. These gating weights control the contribution of each modality to the final fused representation. Subsequently, spatial attention weights are derived from the text-informed multimodal visual features $F_{r}$, using a Multilayer Perceptron (MLP) for feature refinement:
\begin{equation} 
\alpha = SA(F_r) = \sigma (\Phi_{SA}(F_r)),
\end{equation}
where $\Phi_{SA}$ is a fusion function implemented as a MLP consisting of a two-layer convolutional block with a ReLU activation. Finally, the spatial attention weight $\alpha$ and gating weights $G_{v}$ and $G_{i}$ are jointly applied to adaptively weight and fuse the infrared and visible features. This fusion process is defined as:
\begin{equation}
F = \alpha \cdot (1 + G_{v}) \cdot (F_{v} + F_{vi}) \\
+ (1 - \alpha) \cdot (1 + G_{i}) \cdot (F_{i} + F_{iv}),
\end{equation}
where $F$ denotes the final fused feature, which is then passed through the decoder to generate the high-quality fused image $I_{f}$.

\section{Experiments}

\subsection{Experimental Setting}

\noindent
\textbf{Implementation Details.} Our method is implemented in PyTorch and trained across four 24GB NVIDIA GeForce RTX 3090 GPUs. For training, we utilize the AdamW optimizer with an initial learning rate of 0.0001, conducting 140 epochs with a batch size of 8. Training samples are randomly cropped into $96 \times 96$ patches. The model architecture consists of two encoders and a decoder, each comprising four transformer blocks.

\noindent
\textbf{Datasets.} Our experiments utilize three publicly available datasets: MSRS~\cite{tang2022piafusion}, RoadScene~\cite{xu2020u2fusion}, and LLVIP~\cite{jia2021llvip}. The model is trained on the MSRS dataset and evaluated on the RoadScene and LLVIP datasets. The MSRS dataset contains 1,083 training pairs of infrared and visible images, and 361 test pairs. For evaluation, we use 221 image pairs from the RoadScene dataset and 347 randomly selected image pairs from the LLVIP dataset to assess the generalization capabilities of the model across different environments.

\noindent
\textbf{Compared Methods.} We compare our method with nine state-of-the-art (SOTA) approaches: TarDAL~\cite{liu2022target}, ReCoNet~\cite{huang2022reconet}, LRRNet~\cite{li2023lrrnet}, MetaFusion~\cite{zhao2023metafusion}, CDDFuse~\cite{zhao2023cddfuse}, DDFM~\cite{zhao2023ddfm}, Text-IF~\cite{yi2024text}, TextFusion~\cite{cheng2025textfusion}, and MMDRFuse~\cite{deng2024mmdrfuse}.

\noindent
\textbf{Evaluation Metrics.} To quantitatively assess the performance of the proposed fusion model, we adopt five widely used metrics that evaluate various aspects of the fused image quality. These include: Entropy (EN)~\cite{ma2019infrared}, which measures the amount of information retained in the fused image; Standard Deviation (SD), which reflects the richness of information and contrast in the fused image; the Sum of the Correlations of Differences (SCD)~\cite{aslantas2015new}, which assesses the similarity between the fused and source images; Visual Information Fidelity (VIF)~\cite{han2013new}, which quantifies the amount of visual information shared between the fused and source images---higher VIF values indicate better alignment with human visual perception; and $ Q^{AB/F} $~\cite{ma2019infrared}, which evaluates the preservation of edge information from the source images. Higher values across these metrics indicate better performance of the image fusion method.  

\subsection{Comparison with SOTA Fusion Methods}

\noindent
\textbf{Qualitative Comparison.}

\begin{table*}[h]
  \centering
  \large
  \renewcommand{\arraystretch}{1.1}
  \resizebox{\textwidth}{!}{
    \begin{tabular}{c|c c c c c|c c c c c|c c c c c}
    \toprule
    \multirow{2}{*}{Methods} & \multicolumn{5}{c|}{\textbf{MSRS Dataset}} & \multicolumn{5}{c|}{\textbf{RoadScene Dataset}} & \multicolumn{5}{c}{\textbf{LLVIP Dataset}} \\
    \cmidrule(lr){2-6}  \cmidrule(lr){7-11}  \cmidrule(lr){12-16}
    \addlinespace[2pt]
    & EN $\uparrow$ & SD $\uparrow$ & SCD $\uparrow$ & VIF $\uparrow$ & $Q^{AB/F}$ $\uparrow$ & EN $\uparrow$ & SD $\uparrow$ & SCD $\uparrow$ & VIF $\uparrow$ & $Q^{AB/F}$ $\uparrow$ & EN $\uparrow$ & SD $\uparrow$ & SCD $\uparrow$ & VIF $\uparrow$ & $Q^{AB/F}$ $\uparrow$ \\
    
    \midrule
    TarDAL & 5.322 & 23.346 & 0.697 & 0.406 & 0.172 & 7.259 & 47.897 & 1.435 & 0.546 & 0.396 & 6.322 & 37.047 & 1.030 & 0.526 & 0.211 \\
    
    ReCoNet & 4.234 & 41.714 & 1.262 & 0.490 & 0.404 & 7.054 & 41.275 & 1.536 & 0.545 & 0.380 & 5.800 & 46.898 & 1.461 & 0.559 & 0.405 \\
    
    LRRNet & 6.192 & 31.758 & 0.791 & 0.542 & 0.454 & 7.132 & 42.436 & 1.569 & 0.494 & 0.352 & 6.381 & 29.189 & 0.869 & 0.547 & 0.420 \\
    
    MetaFusion & 6.357 & 39.133 & 1.502 & 0.686 & 0.464 & \underline{7.391} & 50.749 & 1.549 & 0.527 & 0.402 & 7.042 & 46.631 & 1.327 & 0.623 & 0.293 \\
    
    CDDFuse & \underline{6.701} & \underline{43.374} & 1.621 & \underline{1.051} & 0.693 & \textbf{7.453} & \textbf{56.322} & 1.712 & 0.624 & 0.483 & 7.342 & 49.971 & \underline{1.580} & 0.880 & 0.642 \\
    
    DDFM & 6.175 & 28.922 & 1.449 & 0.743 & 0.474 & 7.250 & 43.188 & \underline{1.713} & 0.571 & 0.407 & 7.089 & 40.374 & 1.417 & 0.648 & 0.258 \\
    
    Text-IF & 6.665 & 43.190 & \underline{1.689} & 1.049 & \underline{0.716} & 7.319 & 49.871 & 1.522 & \underline{0.686} & \textbf{0.591} & \textbf{7.428} & \underline{50.250} & 1.444 & \underline{1.033} & \underline{0.747} \\
    
    TextFusion & 6.029 & 38.022 & 1.432 & 0.722 & 0.521 & 7.004 & 39.752 & 1.537 & 0.677 & 0.397 & 6.552 & 38.248 & 1.278 & 0.689 & 0.493 \\
    
    MMDRFuse & 6.475 & 37.323 & 1.532 & 0.851 & 0.591 & 6.505 & 27.631 & 1.102 & 0.545 & 0.320 & 7.245 & 44.390 & 1.475 & 0.812 & 0.586\\
    
    \midrule
    \textbf{Ours} & \textbf{6.712} & \textbf{43.472} & \textbf{1.719} & \textbf{1.080} & \textbf{0.733} & 7.180 & \underline{53.678} & \textbf{1.741} & \textbf{0.690} & \underline{0.577} & \underline{7.396} & \textbf{52.553} & \textbf{1.614} & \textbf{1.059} & \textbf{0.763} \\
    
    \bottomrule
    \end{tabular}
  }
  \caption{Quantitative comparison of our method against nine state-of-the-art (SOTA) methods on the MSRS, LLVIP, and RoadScene datasets. The bolded values indicate the best performance, while underlined values represent the second-best.}
  \label{Experi:Tab.1}
\end{table*}

We present a qualitative comparative analysis of the proposed method against nine state-of-the-art (SOTA) fusion methods on the MSRS~\cite{tang2022piafusion}, RoadScene~\cite{xu2020u2fusion}, and LLVIP~\cite{jia2021llvip} datasets, as shown in Fig.~\ref{Experi:Fig.4}. 
The results highlight notable limitations in existing methods. TarDAL, ReCoNet, and LRRNet fail to adequately preserve both infrared and visible information, causing significant information loss (\textit{e.g.}, pedestrians marked by red boxes in MSRS and LLVIP datasets appear indistinct). LRRNet retains some texture details, but TarDAL and ReCoNet struggle with these details (\textit{e.g.}, the ground text marked by red boxes in RoadScene dataset appears blurred). 
TextFusion and MetaFusion show better retention of image details but still have limitations. TextFusion fails to clearly highlight thermal targets (\textit{e.g.}, thermal targets in red boxes are not distinct), and MetaFusion shows noticeable color distortion, with both methods producing images of lower brightness. 
MMDRFuse and DDFM produce low-contrast images, where the overall colors resemble single-modality images, failing to integrate infrared and visible information effectively and making thermal targets less prominent.
CDDFuse and Text-IF generate relatively higher-quality images, preserving infrared information well. However, they blur key visible image texture details, with Text-IF occasionally introducing color distortion (\textit{e.g.}, the background marked by blue boxes in MSRS dataset). 
In contrast, our proposed TeSG method, with its effective textual semantic guidance and gating mechanism, significantly enhances foreground thermal targets while better preserving visible texture details and achieving more realistic brightness. It also presents higher contrast and maintains good contour information and scene details, even under low-light conditions. Our method achieves a balanced fusion of infrared and visible information, delivering higher-quality fusion results both during the day and at night.

\noindent
\textbf{Quantitative Comparison.}
Tab.~\ref{Experi:Tab.1} presents the quantitative evaluation results across three datasets using five metrics. Our method outperforms others on MSRS and LLVIP datasets, particularly excelling on MSRS, where it achieves the highest scores across all metrics. The optimal SD, SCD, and VIF scores indicate that our fused images preserve richer structural and texture information from the source images, demonstrating superior fusion effects in both color and detail. This suggests that our model not only enhances semantic consistency in target regions but also ensures the retention of key features from each modality, while maintaining the overall information richness. Notably, the $Q^{AB/F}$ score shows that our method effectively preserves complementary multimodal features, particularly in retaining edge information, which enhances contrast and contour clarity. Additionally, our method achieves the second-best result in EN, further validating its ability to integrate multi-source information. Overall, our proposed TeSG delivers exceptional performance across multiple datasets, highlighting its strong generalization ability and superior image fusion quality.

\begin{figure*}[!t]
  \centering
  \includegraphics[width=1.0\textwidth]{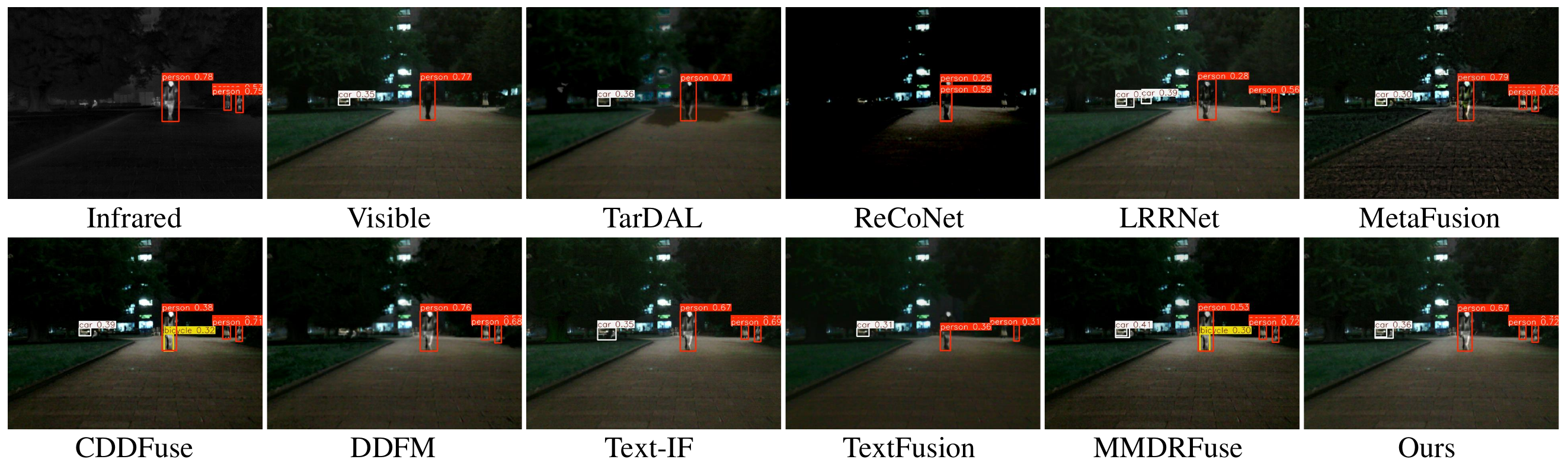}
  \caption{Qualitative comparison of object detection performance on the MSRS dataset.}
  \label{Experi:Fig.5}
\end{figure*}

\subsection{Performance on High-level Tasks}

To comprehensively evaluate the performance of the fused images on downstream high-level vision tasks, we conduct a systematic analysis of various fusion methods on the MSRS dataset~\cite{tang2022piafusion} for both object detection and semantic segmentation.

\noindent
\textbf{Evaluation on the Object Detection Task.}
YOLOv8 \footnote{https://github.com/ultralytics/ultralytics} is employed as the object detection network to ensure a fair qualitative and quantitative comparison. The results are shown in Fig.~\ref{Experi:Fig.5} and Tab.~\ref{Experi:Tab.2}.

The qualitative results in Fig.~\ref{Experi:Fig.5} demonstrate that our method accurately detects all targets in the scene, significantly improving object detection performance and outperforming existing methods. Compared to others, TeSG shows a clear advantage in detection accuracy. For instance, CDDFuse and MMDRFuse incorrectly classify ``person'' as ``bike'', while TextFusion struggles with recognizing the ``person'' category, highlighting its limitations in preserving key thermal features. Additionally, methods like TarDAL, ReCoNet, and DDFM suffer from missed detections due to their inability to capture critical image details, resulting in lower detection performance. In contrast, our method effectively integrates the infrared and visible information, adequately preserving essential features from both modalities. By leveraging the rich semantic information, our method enhances the model's ability to understand and identify scene targets. In challenging scenarios with complex backgrounds, it delivers clearer target contours and richer contextual information, improving detection accuracy and stability.

For quantitative evaluation, we use five metrics, including precision, recall, and mean average precision (mAP) at different thresholds. As shown in Tab.~\ref{Experi:Tab.2}, our method achieves the highest mAP score, notably outperforming others. Even under rigorous metrics such as mAP@0.75 and mAP@0.5:0.95, TeSG demonstrates superior performance in high-precision object detection tasks. These results highlight our method’s superior performance and robustness under stringent detection requirements. 

%
\begin{table}
\centering
\large
\renewcommand{\arraystretch}{1.05}
\resizebox{\columnwidth}{!}{
  \begin{tabular}{c |  c c c c c }
    \toprule
    \multirow{2}{*}{\centering Methods} & \multicolumn{3}{c}{mAP $\uparrow$} & \multirow{2}{*}{\centering Precision $\uparrow$} & \multirow{2}{*}{\centering Recall $\uparrow$} \\ 
    \cmidrule(lr){2-4}     & @0.50 & @0.75 &@0.50:0.95 & & \\ 
    \midrule
    TarDAL           & 0.850 & 0.597 & 0.553 & 0.891 & 0.744 \\
    ReCoNet          & 0.715 & 0.526 & 0.466 & 0.914 & 0.622 \\
    LRRNet           & 0.921 & 0.735 & 0.674 & \textbf{0.971} & 0.852 \\
    MetaFusion       & 0.905 & 0.728 & 0.647 & 0.905 & 0.855 \\
    CDDFuse          & \underline{0.942} & 0.778 & 0.684 & 0.895 & \textbf{0.881} \\
    DDFM             & 0.923 & \underline{0.818} & 0.699 & 0.940 & 0.835 \\
    Text-IF          & 0.939 & 0.808 & \underline{0.709} & 0.947 & 0.860 \\
    TextFusion       & 0.861 & 0.636 & 0.598 & 0.888 & 0.748 \\
    MMDRFuse         & 0.914 & 0.782 & 0.688 & 0.915 & 0.820 \\
    \midrule
    \textbf{Ours}    & \textbf{0.945} & \textbf{0.820} & \textbf{0.717} & \underline{0.948} & \underline{0.879} \\
    \bottomrule
  \end{tabular}
  }
  \caption{Quantitative comparison of object detection performance on the MSRS dataset. The bolded values indicate the best performance, while underlined values represent the second-best.}
  \label{Experi:Tab.2}
\end{table}

\begin{figure*}[!t]
  \centering
  \includegraphics[width=0.9\textwidth]{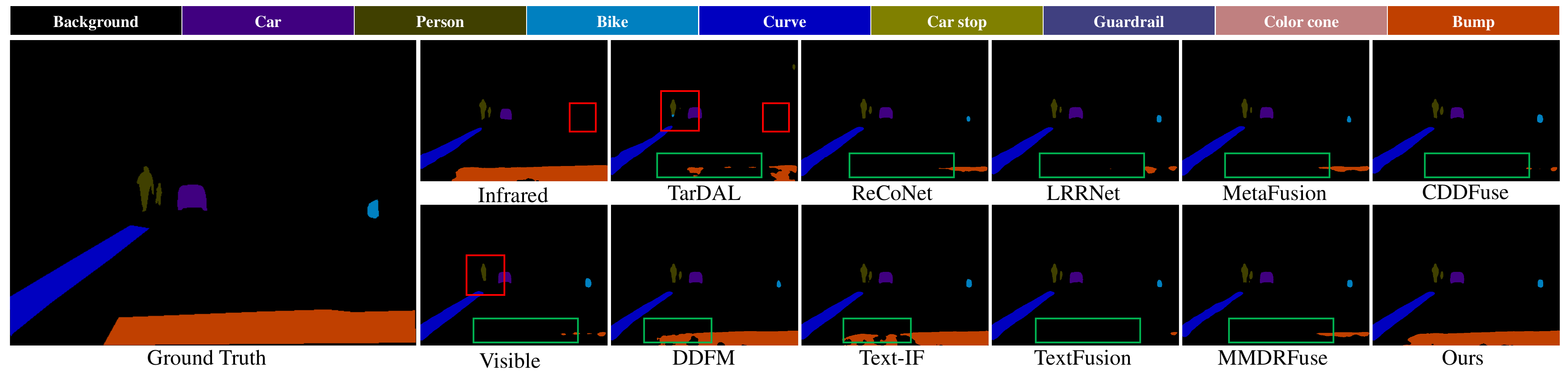}
  \caption{Qualitative comparison of semantic segmentation performance on the MSRS dataset.}
  \label{Experi:Fig.6}
\end{figure*}

\noindent
\textbf{Evaluation on the Semantic Segmentation Task.}
To further evaluate the performance of our method in semantic segmentation, we select DeeplabV3+~\cite{chen2018encoder} as the backbone network and train it on the MSRS dataset, which includes nine categories: background, car, person, bike, curve, car stop, guardrail, color cone, and bump. The model is trained using CrossEntropyLoss with an initial learning rate of 0.01 for 370 epochs and a batch size of 4. The qualitative and quantitative results are presented in Fig.~\ref{Experi:Fig.6} and Tab.~\ref{Experi:Tab.3}.


The qualitative results shown in Fig.~\ref{Experi:Fig.6} indicate that our method accurately segments all semantic targets in the scene, delivering superior performance. In contrast, TarDAL performs poorly in detecting distant targets (\textit{e.g.}, the person and bike marked by red boxes), failing to segment these objects correctly. Methods like ReCoNet, LRRNet, MetaFusion, CDDFuse, TextFusion, and MMDRFuse fail to retain sufficient semantic information in the bump area of the fused images (\textit{e.g.}, the bump marked by green boxes), resulting in its missegmentation. While DDFM and Text-IF successfully identify and segment all targets, they exhibit blurred and imprecise boundaries due to inaccurate edge segmentation. In comparison, our method excels by maintaining consistent edge gradient information. By optimizing edge details in the fusion image, we ensure the edges are highly consistent with those in the input images, enhancing the contrast and clarity of the contours. This significantly improves the performance of our method in downstream tasks like semantic segmentation, further enhancing its practical applicability.

In the quantitative evaluation in Tab.~\ref{Experi:Tab.3}, our method achieves the highest mIoU scores. Our approach outperforms other methods across almost all categories, underscoring its superiority in image segmentation tasks. Specifically, our method effectively emphasizes foreground objects while preserving crucial texture details, giving it a strong advantage in both fusion quality and semantic consistency.

\subsection{Ablation Study}
To assess the contribution of each module to the model's overall performance, we perform ablation experiments on LLVIP~\cite{jia2021llvip} dataset, with the quantitative results provided in Tab. \ref{Experi:Tab.4}.

\noindent
\textbf{Effectiveness of the Mask-Guided Cross-Attention (MGCA) Module.} To validate the effectiveness of the MGCA module, we set up a variant of the module that no longer uses mask semantics to decompose the input images into key target foreground and background regions. Instead, it directly performs feature reconstruction of the entire image using cross-attention. As shown in Tab.~\ref{Experi:Tab.4} (a), most metrics exhibit a decline, while VIF shows a slight increase; however, this improvement is marginal and insufficient to compensate for the overall performance drop. 
This suggests that mask semantics play a crucial role in the fusion process of TeSG, especially for guiding the fusion and preserving image contrast and edge information. 
The drop in SD and SCD further indicates that the masks effectively reduce information loss and improving fusion quality.

\begin{table*}[ht]
  \centering
  \resizebox{0.9 \textwidth}{!}{
    \begin{tabular}{c|ccccccccc c}
      \toprule
                                & \multicolumn{10}{c}{IoU $\uparrow$}                       \\
                                \cline{2-11}
                                \addlinespace[2pt]
      \multirow{-2}{*}{Methods} & Background  &  Car &  Person & Bike  & Curve   & Car stop  & Guardrail  & Color cone     & Bump           & mIoU           \\
      \midrule
      Infrared       & 96.57  & 67.30  & 70.93  & 44.85  & 34.75  & 22.98  &  0.00  & 33.97  & 46.61  & 46.44              \\
      Visible        & 97.91  & 88.03  & 42.00  & 69.55  & 52.39  & 70.30  & {\underline{76.72}}      & 60.56  & 66.05  & 69.28     \\
      TarDAL         & 97.70  & 85.05  & 57.67  & 64.41  & 39.52  & 63.70  & 34.32  & 54.66  & 55.98  & 61.44              \\
      ReCoNet        & 97.55  & 83.95  & 56.54  & 58.69  & 35.54  & 59.52  & 69.24  & 45.33  & 46.03  & 61.38              \\
      LRRNet         & 98.29  & 89.60  & 68.34  & 69.06  & 49.64  & 71.14  & 76.51  & 61.17  & 64.82  & 72.06              \\
      MetaFusion     & 98.15  & 87.92  & 65.25  & 69.18  & 54.51  & 68.18  & 74.61  & 60.09  & 53.43  & 70.15              \\
      CDDFuse        & 98.47  & 90.10  & 74.13  & 72.00  & {\underline{59.85}}      & 71.72  & 76.16  & 61.58  & 65.54  & 74.40     \\
      DDFM           & 98.39  & 89.85  & 73.83  & 70.73  & 56.30  & 70.88  & 72.42  & 60.50  & 63.82  & 72.97              \\
      Text-IF        & {\underline{98.49}}      & {\underline{90.16}}      & {\underline{74.60}}      & {\underline{72.12}}     
                      & 57.89  & 71.85  & \textbf{76.98}  & {\underline{62.59}}      & \textbf{70.38}  & {\underline{75.01}}      \\
      TextFusion     & 98.27  & 89.17  & 65.97  & 70.18  & 54.92  & 71.04  & 75.04  & 59.61  & 64.28  & 72.05              \\
      MMDRFuse       & 98.43  & 89.97  & 73.83  & 71.50  & 58.05  & {\underline{72.04}}      & 74.96  & 61.87  & 64.14  & 73.87     \\
      \midrule
      \textbf{Ours}  & \textbf{98.53}  & \textbf{90.55}  & \textbf{74.86}  & \textbf{72.27}  & \textbf{60.70}  & \textbf{72.42}      
                      & 76.60           & \textbf{62.62}  & {\underline{69.86}}       & \textbf{75.38}           \\
      \bottomrule
    \end{tabular}
    }
  \caption{Quantitative comparison of semantic segmentation performance on the MSRS dataset. The bolded values indicate the best performance, while underlined values represent the second-best.}
  \label{Experi:Tab.3}
\end{table*}

\begin{table}
  \centering
  \resizebox{\columnwidth}{!}{
  \begin{tabular}{l c c c c c }
    \toprule
    Setting  & EN $\uparrow$ & SD $\uparrow$ & SCD $\uparrow$ & VIF $\uparrow$ & $Q^{AB/F}$ $\uparrow$ \\
    \midrule
    (a) w/o MGCA  & 7.366   & 51.690   & 1.576   & \textbf{1.065}   & \textbf{0.765}  \\
    (b) w/o TIVR  & 7.382   & 52.201   & 1.608   & 1.063   & 0.764  \\
    (c) w/o GAF   & 7.383   & 51.956   & 1.572   & 1.065   & 0.764  \\
    \midrule
    (d) \textbf{Ours}  & \textbf{7.396} & \textbf{52.553} & \textbf{1.614} & 1.059 & 0.763  \\
    \bottomrule
  \end{tabular}
  }
  \caption{Quantitative ablation results on LLVIP dataset. The \textbf{bolded} values indicate the best performance.}
  \label{Experi:Tab.4}
\end{table}

\noindent
\textbf{Effectiveness of the Text-Informed Visual Reconstruction (TIVR) Module.} To verify the role of text semantics, we remove the TIVR module, eliminating the guidance of text features, and directly fuse the concatenated image features. As Tab.~\ref{Experi:Tab.4} (b) shows, without text semantic guidance, the fused images exhibit noticeably poorer image information and texture details. This indicates that, in the absence of text guidance, the model struggles to effectively integrate crucial information from the images. Consequently, the fusion lacks detail, particularly in texture retention and semantic consistency. This experiment highlights the significance of text semantic guidance in enhancing the quality of the fused images.

\noindent
\textbf{Effectiveness of the Gated Attentional Fusion (GAF) Module.} To investigate the impact of excluding the gating mechanism for dynamically learning multimodal features on fusion performance, we remove the GAF module and directly decode the initial fused features. As shown in Tab.~\ref{Experi:Tab.4} (c), this resulted in a significant decline in most evaluation metrics, emphasizing the importance of the GAF module. The GAF module dynamically adjusts the fusion weights of different modal features, effectively preserving rich information from both infrared and visible images while maintaining similarity to the source images, thereby significantly improving fusion quality.

\section{Conclusion}

In this paper, we propose TeSG, a novel method for infrared and visible image fusion that integrates textual semantic information. The method includes three core components: the Semantic Information Generator (SIG), the Mask-Guided Cross-Attention (MGCA) module, and the Text-Driven Attentional Fusion (TDAF) module. These components guide the fusion process to fully utilize both mask and text semantics, resulting in enhanced fused images. Extensive experimental results demonstrate the effectiveness of TeSG, showing significant improvements over existing state-of-the-art methods, particularly in terms of performance on downstream tasks such as object detection and semantic segmentation. Future work will explore more controllable fusion mechanisms and broader applications of cross-modal information.

{
    \small
    \bibliographystyle{ieeenat_fullname}
    \bibliography{main}
}


\end{document}